\documentclass[letterpaper, 10 pt, conference]{ieeeconf}  %

\IEEEoverridecommandlockouts                              %
\usepackage{amsmath}
\usepackage{graphicx}
\usepackage{verbatim}
\usepackage{algorithmic}
\usepackage{algorithm}
\usepackage{url}
\usepackage{subcaption}
\usepackage{color}
\usepackage{amsmath}
\usepackage{amssymb}
\usepackage{graphicx}
\usepackage{comment,xspace}
\usepackage{fancybox}
\usepackage{xspace}

\newcommand{\algorlong}{\mathrm{\texttt{Convex Elastic Smoothing}}}
\newcommand{\algor}{\mathrm{\texttt{CES}}}

\newcommand{\rmin}{R_{\mathrm{min}}}
\newcommand{\ulon}{u^{\mathrm{long}}}
\newcommand{\ulat}{u^{\mathrm{lat}}}
\newcommand{\alat}{\mathbf{a}^{\mathrm{lat}}}
\newcommand{\Ulon}{\bar U^{\mathrm{long}}}
\newcommand{\real}{{\mathbb{R}}}
\newcommand{\reals}{\real}

\let\originalc\c
\let\originalv\v
\renewcommand{\c}{\mathbf{c}}

\renewcommand{\v}{\mathbf{v}}

\makeatletter
\providecommand*{\dif}%
        {\@ifnextchar^{\DIfF}{\DIfF^{}}}
\def\DIfF^#1{%
        \mathop{\mathrm{\mathstrut d}}%
                \nolimits^{#1}\gobblespace
}
\def\gobblespace{%
        \futurelet\diffarg\opspace}
\def\opspace{%
        \let\DiffSpace\!%
        \ifx\diffarg(%
                \let\DiffSpace\relax
        \else
                \ifx\diffarg\[%
                        \let\DiffSpace\relax
                \else
                        \ifx\diffarg\{%
                                \let\DiffSpace\relax
                        \fi\fi\fi\DiffSpace}
\makeatother

\renewcommand{\baselinestretch}{0.95}

\title{A Convex Optimization Approach to Smooth Trajectories\\ for Motion Planning with Car-Like Robots\vspace{-0.3cm}}
\author{Zhijie Zhu
    \thanks{Zhijie Zhu is with the Department of Mechanical Engineering, Stanford University, Stanford, CA, 94305, \texttt{zhuzj@stanford.edu}.}
    \and Edward Schmerling
    \thanks{Edward Schmerling is with the Institute for Computational \& Mathematical Engineering, Stanford University, Stanford, CA 94305, \texttt{schmrlng@stanford.edu}.}
    \and Marco Pavone
    \thanks{Marco Pavone is with the Department\ of Aeronautics and Astronautics, Stanford University, Stanford, CA 94305, \texttt{pavone@stanford.edu}.}
}

\graphicspath{{./fig/}}

\begin{document}

\maketitle
\thispagestyle{empty}
\pagestyle{empty}

\begin{abstract}
In the recent past, several sampling-based algorithms have been proposed to compute trajectories that are collision-free and dynamically-feasible. However, the outputs of such algorithms are notoriously jagged. In this paper, by focusing on robots with car-like dynamics, we present a fast and simple heuristic algorithm,  named $\algorlong$ ($\algor$) algorithm, for trajectory smoothing and speed optimization. The $\algor$ algorithm is inspired by earlier work on elastic band planning and iteratively performs shape and speed optimization. The key feature of the algorithm is that both optimization problems can be solved via convex programming, making $\algor$ particularly fast. A range of numerical experiments show that  the $\algor$ algorithm returns high-quality solutions in a matter of a few hundreds of milliseconds and hence appears amenable to a real-time implementation.

\end{abstract}

\section{Introduction}

The problem of planning a collision-free and dynamically-feasible trajectory is fundamental in robotics, with application to systems as diverse as ground, aerial, and space vehicles, surgical robots, and robotic manipulators \cite{SML:06}. A common strategy is to decompose the problem  in steps of computing a collision-free, but possibly highly-suboptimal or not even dynamically-feasible trajectory, smoothing it, and finally reparameterizing the trajectory so that the robot can execute it \cite{SML:11b}. In other words, the first step provides a strategy that explores the configuration space efficiently and decides ``where to go,'' while the subsequent steps provide a refined solution that specifies  ``how to go.''

The first step is often accomplished by running a sampling-based motion planning algorithm \cite{SML:06}, such as PRM \cite{LEK-PS-JCL-MHO:96} or RRT \cite{SML-JJK:01}. While these algorithms are very effective for quickly finding collision-free trajectories in obstacle-cluttered environments, they often return jerky, unnatural paths  \cite{JP-LZ-DM-ea:12}. Furthermore, sampling-based algorithms can only handle rather simplified dynamic models, due to the complexity of exploring the state space while retaining dynamic feasibility of the trajectories. The end result is that the trajectory returned by sampling-based algorithms are characterized by jaggedness and are often dynamically-infeasible, which requires the subsequent use of algorithms for trajectory  smoothing and reparametrization.

Accordingly, the objective of this paper is to design a fast and simple \emph{heuristic} algorithm for trajectory smoothing  and reparametrization that is amenable to a real-time implementation, with a focus on mobile robots, in particular robotic cars. Specifically, we seek an algorithm that within a few hundreds of milliseconds can turn a jerky trajectory returned by a sampling-based motion planner into a smooth, speed-optimized trajectory that fulfills strict dynamical constraints such as friction, bounded acceleration, or turning radius limitations.

\subsection{Literature Review}
 The problem of smoothing a trajectory returned by a sampling-based planner is not new and has been studied since the introduction of sampling-based algorithms \cite{DH:00}. For planning problems that  do not involve the fulfillment of dynamic constraints (e.g., limited turning radius), efficient smoothing algorithms are already available.  In this case, the most widely applied method is the \texttt{Shortcut} heuristic, because of its effectiveness and simple implementation \cite{RG-MHO:07}. In a typical implementation, this algorithm considers two random configurations along the trajectory. If these two configurations can be connected with a new shorter trajectory (as computed via a local planner), then the original connection is replaced with the new one.

For planning problems with dynamic constraints, however, the situation is more contentious. While several works have studied sampling-based algorithms for kinodynamic planning \cite{SML-JJK:01,SK-EF:10,AP-RP-GK-LK-TLP:12,GG-AP-RP-GK:13,SK-EF:13,DW-JvdB:13}, relatively few works have addressed the issues of trajectory smoothing with dynamic constraints. Broadly speaking, current techniques can be classified into two categories \cite{JP-LZ-DM-ea:12}, namely shortcut methods and optimization-based methods. Shortcut methods strive to emulate the \texttt{Shortcut} algorithm in a kinodynamic context. Specifically, jerky portions of a path are replaced with curve segments such as parabolic arcs \cite{PJ-JC:89}, clothoids \cite{SF-PS-JPL-RC:95}, B\'ezier curves \cite{KY-SS:10}, Catmull-Rom splines \cite{SL-XH:11}, cubic B-splines \cite{JP-LZ-DM-ea:12}, or Dubins curves \cite{FL-JPL:01}. Such methods are rather fast (they usually complete in a few  seconds), but handle dynamic constraints only implicitly, for example, by constraining the curvature of the trajectories and/or ensuring $C^2$ continuity. Also, they usually do not involve speed optimization along the computed trajectory.
Notably, two of the main teams in the DARPA Grand Challenge, namely  team Stanford \cite{ST-MM:06} and team CMU \cite{CU-JA-ea:07},  applied shortcut methods as their smoothing procedure.

In contrast, optimization-based methods handle dynamic constraints explicitly, as needed, for example, for high-performance mobile vehicles. Two common approaches are gradient-based methods \cite{NR-MZ-JAB-ea:09} and elastic bands or elastic strip planning  \cite{JB-WH:08,OB-OK:02,SQ-OK:93}, which model a trajectory as an elastic band. These works, however, are mostly geared toward robotic manipulators \cite{OB-OK:02}, may still require several seconds to find a solution \cite{NR-MZ-JAB-ea:09}, and generally do not address speed optimization along the trajectory.

\subsection{Statement of Contributions}
In this paper, leveraging recent strides  in the field of convex optimization \cite{SB-LV:04}, we design a novel algorithm, named $\algorlong$ ($\algor$) algorithm, for trajectory smoothing and speed optimization.  The focus is on mobile robots with car-like dynamics.  Our algorithm is inspired by the elastic band approach \cite{SQ-OK:93}, in that we identify a collision-free ``tube" around the trajectory returned by a sampling-based planner, within which the trajectory is ``stretched" and speed is optimized. The stretching and speed optimization steps rely on convex optimization. In particular, the stretching step draws inspiration from \cite{SE-SF-JCG:13}, while the speed optimization step is essentially an implementation of the algorithm in \cite{TP-SB:14}. In contrast with \cite{SQ-OK:93}, our algorithm uses convex optimization for the stretching process, performs speed optimization, and handles a variety of constraints (e.g., friction) that were not considered in  \cite{SQ-OK:93}. As compared to \cite{SE-SF-JCG:13}, our algorithm handles more general workspaces and removes the assumption of a constant speed.

Specifically, the $\algor$ algorithm divides  the trajectory smoothing process into two steps: (1) given a fixed velocity profile along a reference trajectory, optimize the shape of the trajectory, and (2) given a fixed shape for the trajectory, optimize the speed profile along the trajectory. We show that each of the two steps can be readily solved as a convex optimization problem. The two steps are then repeated until a termination criterion is met (e.g., timeout). In this paper, the initial reference trajectory is computed by running the differential FMT$^*$ algorithm \cite{ES-LJ-MP:15a}, a kinodynamic variant of the FMT$^*$ algorithm \cite{LJ-ES-AC-ea:15}. Numerical experiments on a variety of scenarios show that in a few \emph{hundreds of milliseconds} the $\algor$ algorithm outputs a ``high-quality" trajectory, where the jaggedness of the original trajectory is eliminated and speed is optimized.  Coupled with differential FMT$^*$, the $\algor$ algorithm is able to find high-quality solutions to rather complicated planning problems in well under a second, which appears promising for a real-time implementation.

We mention that the reference trajectory used as an input to the $\algor$ algorithm can be the output of any sampling-based motion planner, and indeed of any motion planning algorithm. In particular, the $\algor$ algorithm  appears to perform well even when the reference trajectory is not collision-free or does not fulfill some of the dynamic constraints (see Section \ref{sec:sim} for more details). Also, while in this paper we mostly focus on vehicles with second-order, car-like dynamics, the $\algor$ algorithm can be generalized to a variety of other mobile systems such as aerial vehicles and spacecraft.

\subsection{Organization}
This paper is structured as follows. In Section \ref{sec:ProbState} we formally state the problem we wish to solve. In Section \ref{sec:algo} we present the $\algor$ algorithm, a novel algorithm for trajectory smoothing that relies on convex optimization and runs in a few hundreds of milliseconds. In Section \ref{sec:sim} we present results from numerical experiments highlighting the speed of the the $\algor$ algorithm and the quality of the returned solutions. Finally, in Section \ref{sec:conc}, we draw some conclusions and discuss directions for future work.

\section{Problem Statement}\label{sec:ProbState}
\begin{figure}[h!]
  \centering
    \includegraphics[width=0.35\textwidth]{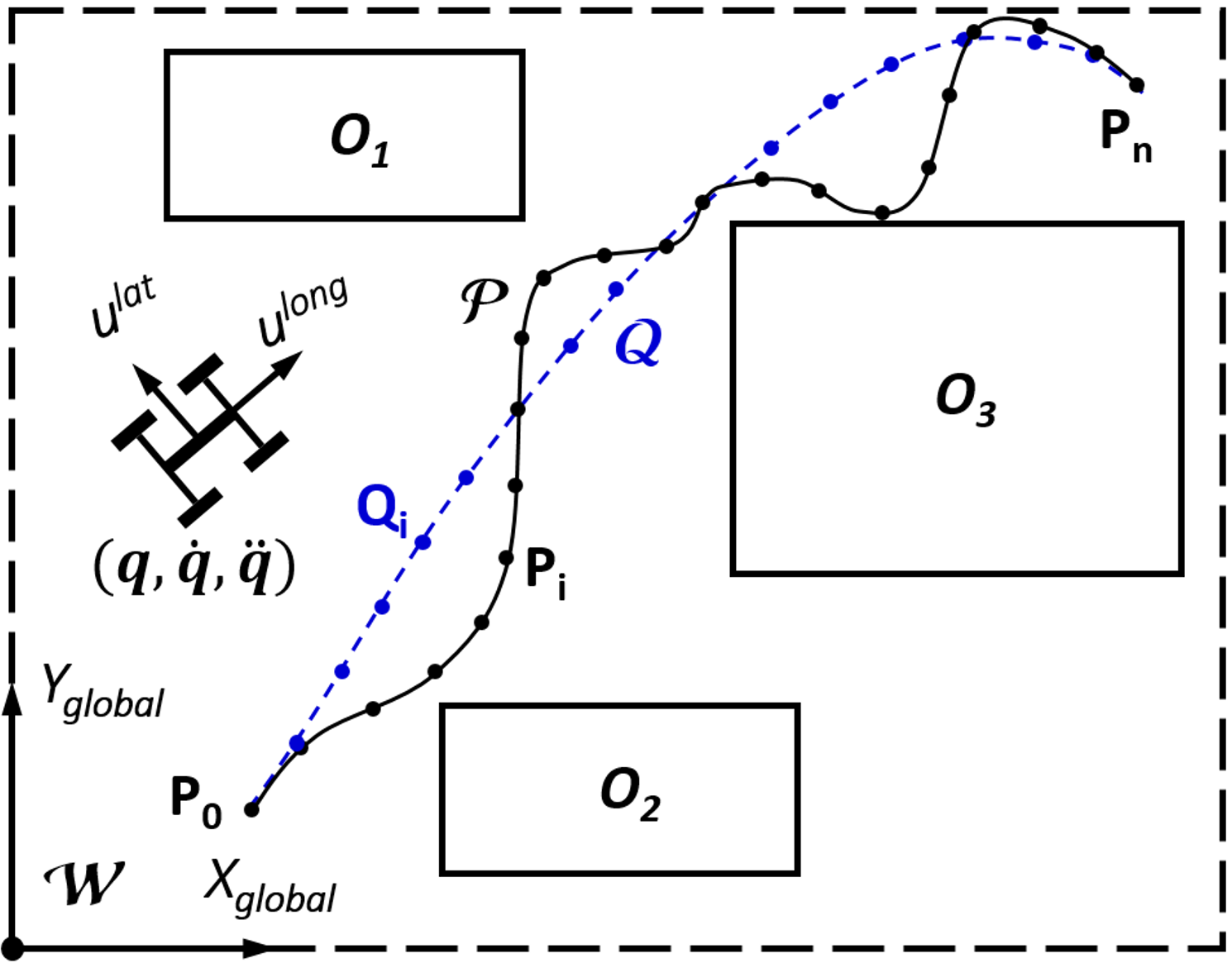}
    \caption{The goal of this paper is to design a fast algorithm to locally optimize the output of a motion planner, with a focus on car models. Specifically, the smoothing algorithm takes as input a reference trajectory $\mathcal P$ and returns a smoothed trajectory $\mathcal Q$ with an optimized speed profile.}
    \label{fig:prob_form}
\end{figure}

Let $\mathcal{W} \subset \reals^2$ denote the two-dimensional  work space for a car-like vehicle.  Let $\mathcal{O}=\{ O_1, O_2,\ldots,O_{m}\}$, with $O_i \subset \mathcal W$, $i=1,\ldots, m$, denote the set of obstacles. For simplicity, we assume that the obstacles have polygonal shape. In this paper we primarily focus on a unicycle dynamic model for the vehicle. Extensions to more sophisticated car-like models are discussed in Section \ref{sec:sim}.
Specifically, following \cite{TP-SB:14}, let $\mathbf{q} \in \mathcal W$ represent the position  of the vehicle, and $\mathbf{\dot{q}}$ and $\mathbf{\ddot{q}}$ its velocity and acceleration, respectively. We consider a non-drifting and non-reversible car model so that the heading of the vehicle is the same as the direction of the instantaneous velocity vector, and we denote by $\phi(\mathbf{\dot{q}}) $ the mapping from vehicle's speed to its heading. The control input $\mathbf{u}  = [\ulon, \ulat]$ is two-dimensional, with the first component, $\ulon$, representing longitudinal force and the second component, $\ulat$, representing lateral force. The dynamics of the vehicle are given by
\begin{equation}\label{eq:dyn}
m\, \mathbf{\ddot{q}}  = \begin{bmatrix}
       \cos \phi(\mathbf{\dot{q}})  & -\sin \phi(\mathbf{\dot{q}}) \\[0.3em]
       \sin \phi(\mathbf{\dot{q}})  & \cos \phi(\mathbf{\dot{q}})
     \end{bmatrix}\, \mathbf{u},
\end{equation}
where $m$ is the vehicle's mass, see Figure \ref{fig:prob_form}. We consider a friction circle constraint for $\mathbf{u}$, namely
\begin{equation}\label{eq:con_1}
\|\mathbf{u}\| \leq \mu \, m \, g,
\end{equation}
where $\mu$ is the friction coefficient and $g$ is the gravitational acceleration (in this paper, norms should be interpreted as 2-norms). The longitudinal force is assumed to be upper bounded as
\begin{equation}\label{eq:con_2}
\ulon \leq \Ulon,
\end{equation}
where $\Ulon \in \reals_{>0}$ encodes the force limit
from the wheel drive.  Finally, we assume a minimum turning radius $\rmin$ for the vehicle, which, in turn, induces a constraint on the lateral force according to
\begin{equation}\label{eq:con_3}
\ulat \leq m \frac{\| \mathbf{\dot q}\|^2}{\rmin}.
\end{equation}
The minimum  turning radius ($\rmin$) depends on specific vehicle parameters such as wheelbase and maximum steering angle for steering wheels.

Some comments are in order. First, the unicycle model assumes that  the heading of the vehicle is the same as the direction of the instantaneous velocity vector. This is a reasonable assumption in most practical situations, but becomes a poor approximation at high speeds when significant under-steering takes place, or at extremely low speeds when the motion is determined by Ackermann steering geometry. We will show, however, that the results presented in this paper can be extended to more complex car models, e.g., half-car models, by leveraging differential flatness of the dynamics. Second, we assume that the actual control inputs such as steering and throttle opening angles can be mapped to $\mathbf{u}$ via a lower-level control algorithm. This is indeed true for most ground vehicles, see, e.g., \cite{DLM:03}. Third,
the friction circle model captures the dependency between lateral and longitudinal forces in order to prevent sliding. Fourth, constraint \eqref{eq:con_2} might yield unbounded speeds. This issue could be addressed by conservatively choosing a smaller value of $\Ulon$ that guarantees an upper bound of the achievable speed within the planning distance. An alternative formulation is to set a limit on the total traction power $\ulon\ \| \mathbf{\dot q}\|  \leq W$, where $W$ denotes the maximum power provided by the engine. This form is closer to the real constraint on traction force, but it makes the constraint non-convex -- this is a topic left for future research. Finally, the  model \eqref{eq:dyn}-\eqref{eq:con_3}, with minor modifications, can be applied to a variety of other vehicles and robotic systems, e.g., spacecraft, robotic manipulators, and aerial vehicles \cite{TP-SB:14}. Hence, the algorithm presented in this paper may be applied to a rather large class of systems -- this is a topic left for future research.

We are now in a position to state the problem we wish to solve in this paper.  Consider a collision-free \emph{reference trajectory} computed, for example,  by running a sampling-based motion planner \cite{SML:06}.  Let this trajectory be discretized into a set of waypoints $\mathcal{P} := \{P_0, P_1,\ldots,P_n\}$, where, by construction,  $P_i \in \mathcal W\setminus \mathcal O$ for $i=1,\ldots, n$. The goal is to design a heuristic \emph{smoothing algorithm} that uses the information about the vehicle's model \eqref{eq:dyn}-\eqref{eq:con_3}, obstacle set $\mathcal{O}$, and (discretized) reference trajectory $\mathcal{P}$ to compute a dynamically-feasible (with respect to model \eqref{eq:dyn}-\eqref{eq:con_3}), collision-free, and smooth trajectory that goes from $P_0$ to $P_n$ and has an optimized speed profile, see Figure \ref{fig:prob_form}. Our proposed algorithm is named $\algor$ and is presented in the next section.

\vspace{-1em}
\section{The $\algor$ Algorithm}\label{sec:algo}
At a high level, the $\algor$ algorithm performs the following operations. First, a sequence of ``bubbles" is placed along the reference trajectory in order to identify a region of the workspace that is collision free. Such a region can be thought of as a collision-free ``tube" within which the reference trajectory, imagined as an elastic band, can be stretched so as to obtain a smoother trajectory. Assuming that a speed profile along the reference trajectory is given, such stretching procedure can be cast as a convex optimization problem, as it will be shown in Section \ref{subsec:es}. Furthermore, optimizing the speed profile along a stretched trajectory can also be cast as a convex optimization problem. However, jointly stretching a trajectory and optimizing the speed profile is a non-convex   problem. Hence,  the $\algor$ algorithm proceeds by alternating trajectory stretching and speed optimization.  Simulation results, presented in Section \ref{sec:sim}, show that such a procedure is amenable to a real-time implementation and yields suboptimal, yet high-quality trajectories. Our $\algor$ algorithm is inspired by the elastic band and bubble method \cite{SQ-OK:93}. The work in \cite{SQ-OK:93}, however, mostly focuses on geometric (i.e., without differential constraints) planning, and does not consider speed optimization, as opposed to our problem setup.

In the remainder of this section we present the different steps of the $\algor$ algorithm, namely, bubble generation, elastic stretching and speed optimization. Finally the overall $\algor$ algorithm is presented.

\subsection{Bubble Generation}\label{subsec:bg}

The first step is to compute a sequence of bubbles, one for each waypoint $P_i$, so as to identify a collision-free tube along the reference trajectory for subsequent optimization. Since the problem is two-dimensional, each bubble is indeed a circle. As discussed, extensions to systems in higher dimensions (e.g., airplanes or quadrotors) are possible, but are left for future research. The bubble generation algorithm is shown in Algorithm \ref{BubbleGeneration}.
\begin{algorithm}[t]
\caption{Bubble generation }
\label{BubbleGeneration}
\algsetup{linenodelimiter=}
\begin{algorithmic}[1]
\REQUIRE Reference trajectory $\mathcal P$, obstacle set $\mathcal O$, bubble bounds $r_l$ and $r_u$
\FOR{ $i = 2:(n -1)$ }
\IF{$\|P_i - A_{i-1}\|<0.5\cdot r_{i-1}$} \label{line:ifstart}
\STATE $\mathcal{B}_{i} \leftarrow \mathcal{B}_{i-1}$
\STATE $A_i \leftarrow A_{i-1}$
\STATE $r_i \leftarrow r_{i-1}$
\STATE $\bf{continue}$
\ENDIF \label{line:ifends}
\STATE $\mathcal{B}_{i} \leftarrow \texttt{GenerateBubble}(P_i)$ \label{line:genstart}
\STATE $r_{i} \leftarrow$ \text{Radius of $\mathcal{B}_{i}$}
\STATE $A_i \leftarrow P_i$ \label{line:genend}
\IF {$r_i < r_l$} \label{line:movestart}
\STATE $\mathcal{B}_{i} \leftarrow \texttt{TranslateBubble}(A_i, r_i)$
\STATE $r_{i} \leftarrow$ \text{Radius of $ \mathcal{B}_{i}$}
\STATE $A_i \leftarrow $ Center of $\mathcal B_i$
\ENDIF \label{line:moveend}
\ENDFOR
\end{algorithmic}
\end{algorithm}

Let $\mathcal B_i$ denote the bubble associated with waypoint $P_i$, $i=1,\ldots, n$, and $A_i$ and $r_i$ denote, respectively, its center and radius. According to this notation, $\mathcal B_i = \{x\in \mathcal W \, | \, \|x-A_i \| \leq r_i\}$. For each waypoint $P_i$, Algorithm  \ref{BubbleGeneration} attempts to compute a bubble such that: (1) its radius is upper bounded by $r_u \in \reals_{>0}$, (2) \emph{whenever possible}, its radius is no less than  $r_l \in \reals_{>0}$, and (3) its center is as close as possible to $P_i$. The role of the upper bound $r_u$ is to limit the smoothing procedure within a relatively small portion of the workspace, say 10\% (in other words, to make the optimization ``local"). In turn, the minimum bubble radius $r_l$ is set according to the maximum distance between adjacent waypoints, so that every bubble overlaps with its neighboring bubbles and the placement of new waypoints for trajectory stretching (see Section \ref{subsec:es}) does not have any ``gaps".

First, in lines \eqref{line:ifstart}-\eqref{line:ifends}, the algorithm checks whether $P_i$ is ``too close" to the center of bubble $\mathcal B_{i-1}$. If this is the case, bubble $\mathcal B_i$ is made equal to bubble $\mathcal B_{i-1}$. Otherwise, the algorithm considers as candidate center for bubble $\mathcal B_i$ the waypoint $P_i$ (that is collision-free) and computes, via function \texttt{GenerateBubble}($P_i$), the largest bubble centered at $P_i$ that is collision-free (with maximum radius $r_u$), see lines \eqref{line:genstart}-\eqref{line:genend}. For rectangular-shaped obstacles, this is a straightforward geometrical procedure. If the obstacles have more general polygonal shapes, a bisection search is  performed with respect to  the bubble radius.
Should the resulting radius be lower than the threshold $r_l$, an attempt is made to translate  $A_i$ so that a larger (collision-free) bubble can be placed, lines \eqref{line:movestart}-\eqref{line:moveend}. Specifically, function $\texttt{TranslateBubble}(A_i, r_i)$ first identifies the edge of the obstacle closest to $A_i$ (recall that the obstacles are assumed of polygonal shape). Then, the outward normal direction to the edge is computed and the center of the bubble is moved along such direction until a ball of radius $r_l$ can be placed. Should this not be possible, then $\texttt{TranslateBubble}(A_i, r_i)$ returns the ball of largest radius among the balls whose centers lie on the aforementioned normal direction.

\begin{figure}[h!]
  \centering
    \includegraphics[width=0.3\textwidth]{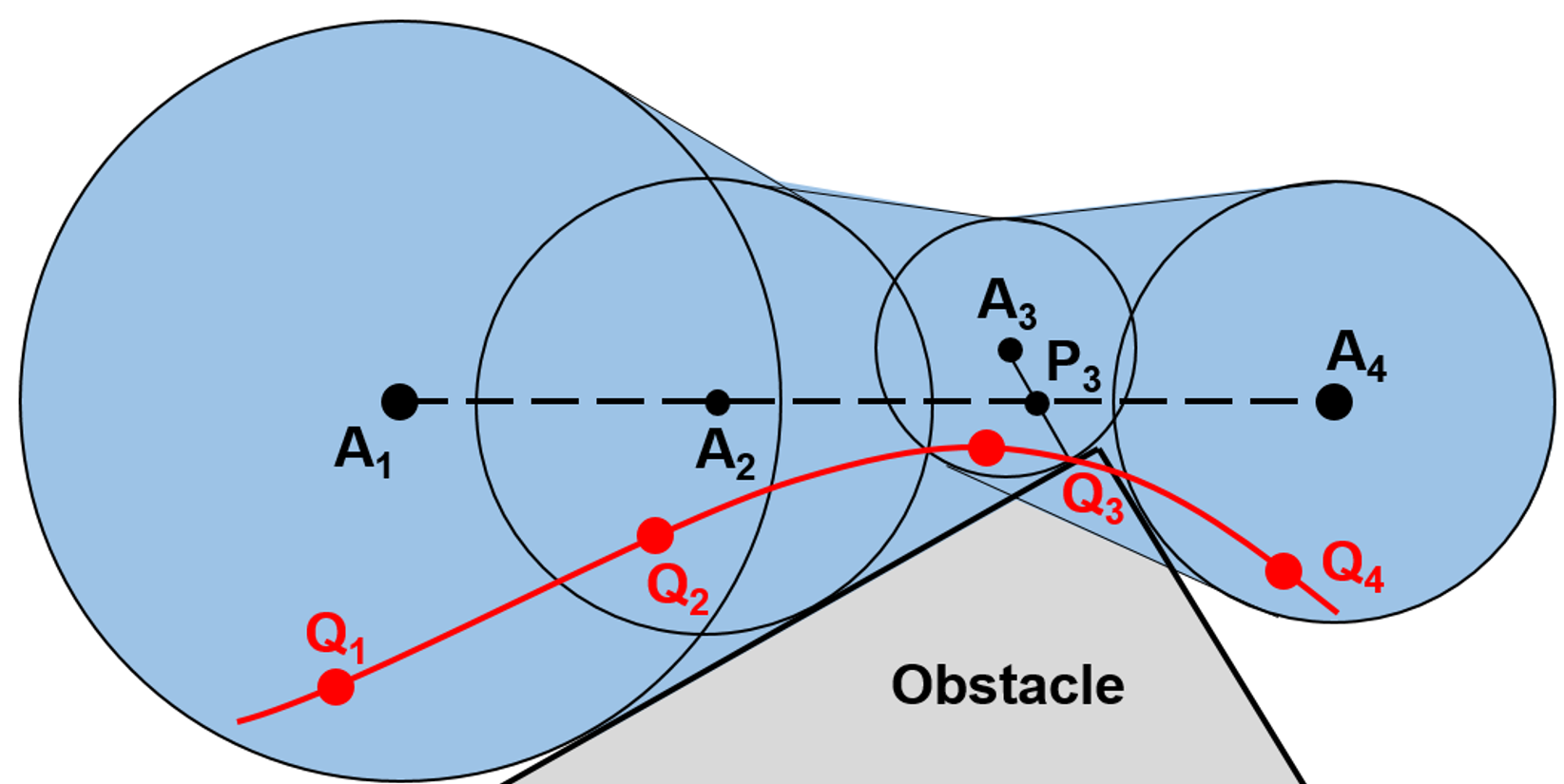}
    \caption{Example application of the bubble generation procedure, Algorithm \ref{BubbleGeneration}.}
    \label{fig:bubble_illu}
\end{figure}

\begin{figure}[h!]
  \centering
    \includegraphics[width=0.3\textwidth]{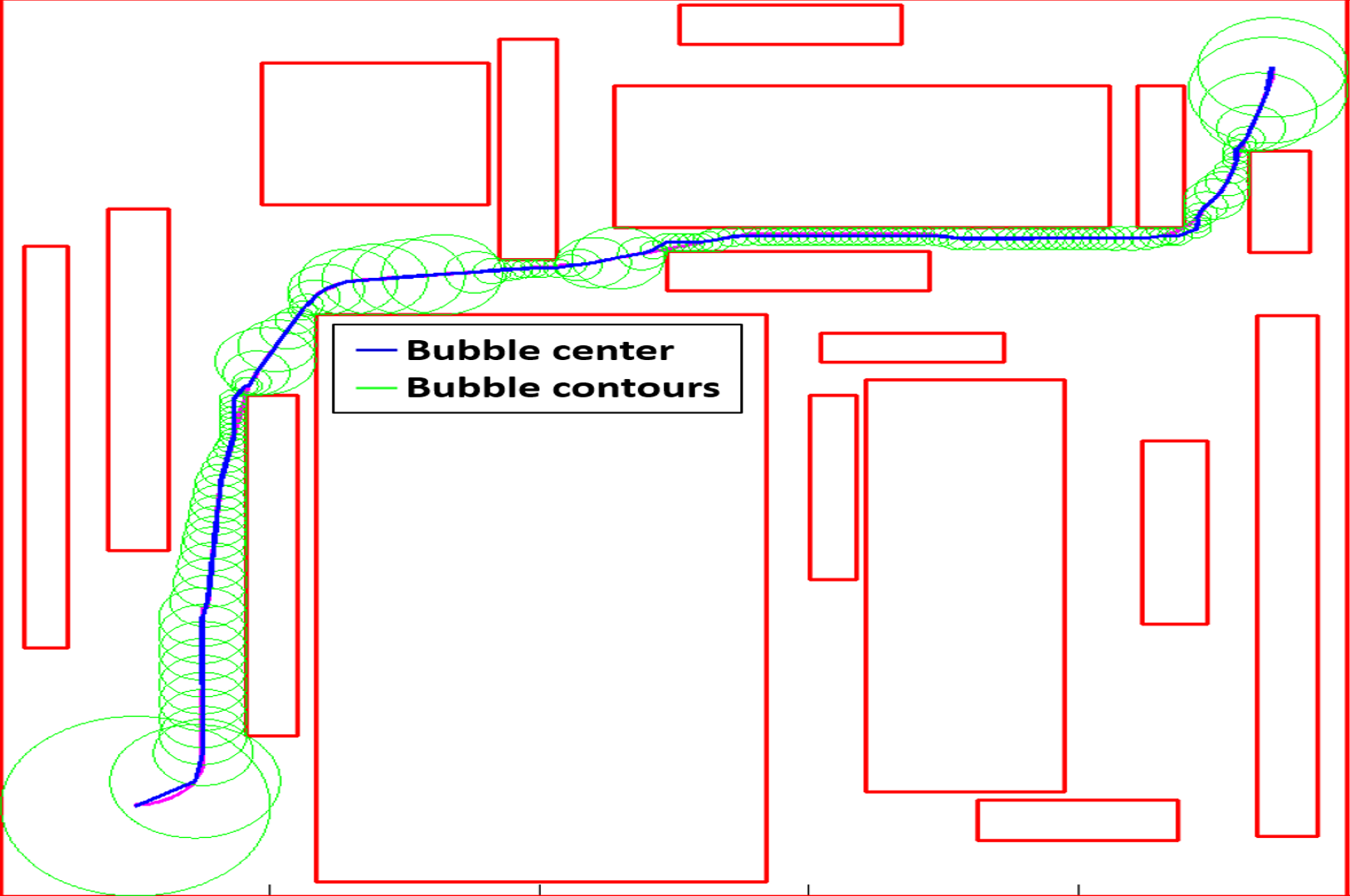}
    \caption{Typical output of Algorithm \ref{BubbleGeneration} for an environment with rectangular-shaped obstacles.}
    \label{fig:bubble_exp}
\end{figure}

Figure \ref{fig:bubble_illu} shows an example application of the bubble generation algorithm. The centers $A_1$, $A_2$ and $A_4$ coincide with their corresponding waypoints, while center  $A_3$ is translated away from waypoint $P_3$ to allow for a larger bubble radius. Figure \ref{fig:bubble_exp} shows a typical output of Algorithm~\ref{BubbleGeneration}.

By construction, the bubble regions are collision-free and represent the feasible space for the placement of new optimized waypoints during the elastic stretching  process. Note that in some cases trajectories connecting points in adjacent bubbles may be  in collision with obstacles, as shown in Figure \ref{fig:bubble_illu}. This issue is mitigated in practice by considering a ``large enough" number of reference waypoints (possibly adding them iteratively) and/or inflating the obstacles. A principled way to select the number of waypoints would rely on a reachability analysis for the unicycle model \eqref{eq:dyn}-\eqref{eq:con_3}. However, to minimize computation time, we rely on a heuristic choice for the waypoint number. Specifically, the spacing between the waypoints is roughly equal to a quarter of the car length.

\vspace{-1.5em}
\subsection{Elastic Stretching (aka Shape Optimization)}\label{subsec:es}
\begin{figure}[h!]
  \centering
    \includegraphics[width=0.28\textwidth]{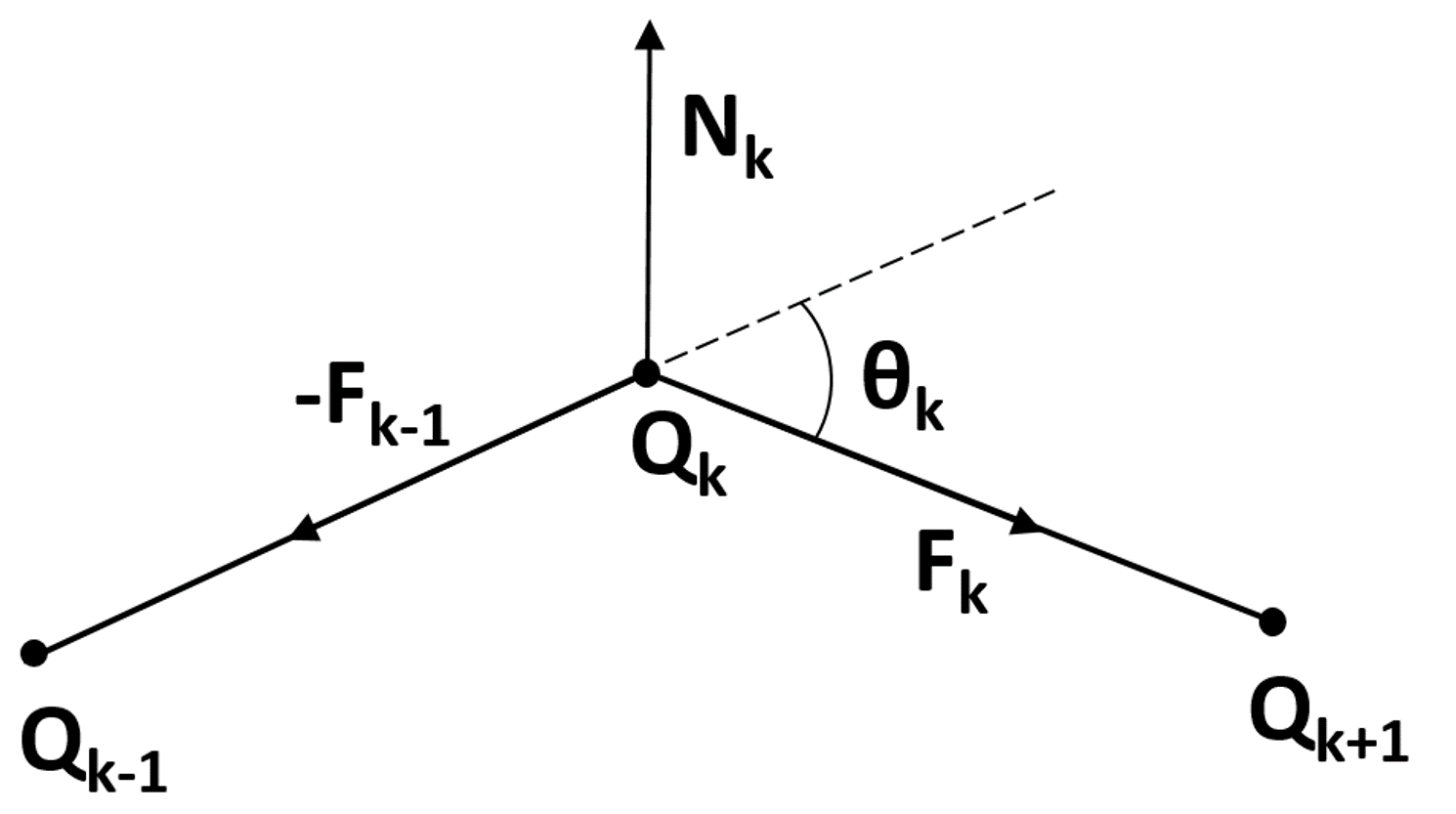}
    \caption{Artificial tensile forces and balance force for a sequence of points placed in bubbles $k-1$, $k$, $k+1$.
    \vspace{-1.5em}}
    \label{fig:elastic_band}
\end{figure}

The key insight of the elastic stretching procedure is to view a trajectory as an elastic band, with $n$ nodal points whose positions can be adjusted within the respective (collision-free) bubbles. Such nodal points represent the new waypoints for the smoothed trajectory, which replace the original waypoints in $\mathcal P$. Within this perspective,  the dynamic constraints on the shape of a trajectory are mimicked by the bending stiffness of the band.

Specifically, consider Figure  \ref{fig:elastic_band}. Let $Q_{k}$ and $Q_{k+1}$ be points, respectively, in bubbles $k$ and $k+1$, with $k=1,\ldots, n-1$. We consider an \emph{artificial} tensile force $\mathbf{F}_k$ between $Q_k$ and $Q_{k+1}$ given by
\[
\vspace{-.5em}
\mathbf{F}_k:= Q_{k+1} - Q_{k}.
\vspace{-.2em}
\]
Accordingly, the balancing force at point $Q_k$, $k=1,\ldots, n$~is
\begin{equation}\label{eq:balanceForce}
\vspace{-.5em}
\mathbf{N}_{k} := \mathbf{F_{k-1}} - \mathbf{F_{k}}.
\end{equation}

As in  \cite{SQ-OK:93}, the physical interpretation is a series of springs between the bubbles. From a geometric standpoint, the balancing force $\mathbf{N}_{k}$  captures curvature information along a trajectory. Note that if all balance forces are equal to zero, the trajectory is a straight line, which, clearly, has ``ideal smoothness."  To smooth a trajectory, the goal is then to place new waypoints $Q_1, \ldots, Q_n$ within the bubbles so as to minimize the sum of the norms of the balance forces subject to constraints due to the vehicle's dynamics. In this way, the trajectory is ``bent'' as little as possible in order to avoid collisions with obstacles.

The constraints for the placement of the new waypoints rely on a  number of approximations  to ensure convexity of the optimization problem.  Specifically, assume the longitudinal force $\ulon_k$ and velocity $\mathbf{v_k}$ are given at each waypoint in $\mathcal P$ (their optimization will be discussed in the next section). We define $R_k$ as the instantaneous turning radius for the car at the $k$th waypoint. The lateral acceleration  $\alat$ for waypoints $Q_k$, $k=2, \ldots, n-1$ can be upper bounded as
\vspace{-.3em}
\begin{equation}\label{eq:normFprime3}
\alat = \frac{\|\mathbf{v_k}\|^2}{R_k} \leq \sqrt{\left (\mu \, g \right )^2 - \left (\frac{\ulon_k}{m} \right)^2} : =  \alpha_k,
\end{equation}
where the inequality follows from the friction circle constraint \eqref{eq:con_1}. Hence, $1/R_k\leq \alpha_k/\|\mathbf{v_k}\|^2$. To relate $R_k$ with $\|\mathbf{N}_{k}\|$, we make use of the following approximations, valid when the waypoints are uniformly spread over a trajectory and dense ``enough'': (1) $\|\mathbf{F}_{k} \| \cong \|\mathbf{F}_{k-1}\|$, (2) $\theta_k$ is small, and (3) $\theta_k\cong \|Q_{k+1} - Q_k \|/R_k$. Then one can write

\begin{equation}\label{eq:approx}
\begin{split}
\|\mathbf{N}_k\| &= \|\mathbf{F}_{k-1} - \mathbf{F}_{k}\| \cong 2 \cdot \|\mathbf{F}_k\| \cdot \sin(\theta_k/2) \\
&\cong \|Q_{k+1} - Q_k \| \, \theta_k \cong \|Q_{k+1} - Q_k \|^2 /R_k\\\
& \leq \|Q_{k+1} - Q_k \|^2 \, \alpha_k/\|\mathbf{v_k}\|^2.
\end{split}
\end{equation}
Lastly, to make the above inequality a quadratic constraint in the $Q_k$'s variables, we approximate the length of each \emph{band} $Q_{k+1}- Q_k$ as the average length $d$ along the reference trajectory, i.e.,%
\[
d: = \frac{\sum_{k=1}^{n-1} \|P_{k+1} - P_{k}\|}{n-1}.
\]

\noindent Note that the minimization of $\|\mathbf{N}_k\|$ as the optimization objective inherently reduces the non-uniformity of the lengths of each \emph{band}, which justifies the above approximation.

In summary, we obtain the \emph{friction} constraint
\begin{equation}\label{eq:frictionconstraint}
\|\mathbf{N}_k\| \leq \alpha_k \, \left (\frac{d}{||\mathbf{v}_k||}\right)^2.
\end{equation}
Also, again leveraging equation \eqref{eq:approx}, we obtain the \emph{turning radius} constraint
\begin{equation}\label{eq:kinematicconstraint}
\|\mathbf{N}_k\| \leq  \frac{d^2}{\rmin}.
\end{equation}

To constrain the initial and final endpoints of the trajectory, one simply imposes $Q_1 = P_1$ and $Q_n = P_n$. In turn, to constrain the initial and final heading angles, one can use the constraints $Q_2 = P_1 + d \cdot \frac{\mathbf{v}_1}{\|\mathbf{v}_1\|} $ and $Q_{n-1} = P_n - d \cdot \frac{\mathbf{v_{n-1}}}{\|\mathbf{v_{n-1}}\|}$. Noting that $\mathbf{N}_k = 2Q_k -Q_{k-1}  -Q_{k+1}$, the elastic stretching optimization problem is:
\vspace{-.2em}
\begin{alignat*}{2}
\min_{Q_3,\ldots, Q_{n-2}} & & \quad& \sum_{k=2}^{n-1} \|2Q_k -Q_{k-1}  -Q_{k+1}\|^2\\
\mathrm{s.t.}\qquad & &\quad & \mkern-36mu  Q_1 = P_1,\ Q_2 = P_1 + d \cdot \frac{\mathbf{v}_1}{\|\mathbf{v}_1\|} \\
& &\!\quad & \mkern-36mu  Q_n = P_n,\ Q_{n-1} = P_n - d \cdot \frac{\mathbf{v}_{n-1}}{\|\mathbf{v}_{n-1}\|}\\
& &\!\quad & \mkern-36mu  Q_{k} \in \mathcal{B}_k, \quad k = 3,\ldots, n-1\\
& &\quad & \mkern-36mu {\small \|2Q_k -Q_{k-1}  -Q_{k+1}\| \leq \min \left\{ \frac{d^2}{\rmin}, \, \alpha_k \left (\frac{d}{\mathbf{v}_k}\right)^2\right\}}\\
& &\quad& \qquad \text{for } k=2,\ldots, n-1.
\end{alignat*}
This is a convex optimization problem with quadratic objective and quadratic constraints (QCQP), where the decision variables are the intermediate waypoints $Q_k$, $k=3, \ldots, n-2$, which can be placed anywhere within  the collision-free regions $\{\mathcal{B}_3,\ldots,\mathcal{B}_{n-2}\}$,

Note that any feasible solution to the above QCQP, output as a sequence of discrete waypoints, represents a continuous-time trajectory satisfying the unicycle model \eqref{eq:dyn}-\eqref{eq:con_3}. This full trajectory may be recovered by interpolating between adjacent waypoints $Q_k$,$Q_{k+1}$ using circular arcs centered at the intersection of $\mathbf{v}_{k}^\perp, \mathbf{v}_{k+1}^\perp$, or a straight line if the two velocity vectors are parallel. The speed profile along this continuous trajectory may be taken as piecewise linear, and the discrete constraints \eqref{eq:frictionconstraint} and \eqref{eq:kinematicconstraint} ensure that the continuous constraints \eqref{eq:con_1}-\eqref{eq:con_3} are satisfied.
The quality of such trajectories will be investigated in Section \ref{sec:sim}.

\subsection{Speed Optimization}\label{subsec:so}

The speed optimization over a fixed trajectory relies on the convex optimization algorithm presented in \cite{TP-SB:14}. The inputs to this algorithm are (1) a sequence of  waypoints $\{Q_1,\ldots,Q_n\}$ (representing the trajectory to be followed), (2) the friction coefficient $\mu$ for the friction circle constraint in equation \eqref{eq:con_1}, and (3) the maximum traction force $\Ulon$ defined in equation \eqref{eq:con_2}. The outputs   are (1) the sequence of velocity vectors $\{\mathbf{v}_1,\ldots,\mathbf{v}_n\}$ and (2) the sequence of longitudinal control forces $\{\ulon_1,\ldots,\ulon_n\}$, one for each waypoint $Q_k$, $k=1,\ldots, n$. We refer the reader to \cite{TP-SB:14} for details about the algorithm.

\subsection{Overall Algorithm}
The $\algor$ algorithm alternates between elastic stretching (Section \ref{subsec:es}) and speed optimization (Section \ref{subsec:so}), until a given tolerance on length reduction, traversal time reduction, or a timeout condition are met. Note that at iteration $i\geq2$, the elastic stretching algorithm should use as estimate for the average band length the quantity
\[
d^{[i]}: = \frac{\sum_{k=1}^{n-1} \left \|Q^{[i-1]}_{k+1} - Q^{[i-1]}_{k}\right \|}{n-1},
\]
where the $Q_k^{[i-1]}$'s are the waypoints computed at iteration $i-1$. According to our discussion in Section \ref{subsec:es}, at iteration $i=1$ one should set $Q_k^{[0]} = P_k$, for $k=1, \ldots, n$.

\section{Numerical Experiments}\label{sec:sim}

In this section we investigate the effectiveness of the $\algor$ algorithm along two main dimensions: (1) quality of the smoothed trajectory, measured in terms of traversal time reduction with respect to the reference trajectory, and (2) computation time. We consider three sets of experiments. In the first set, we consider 24 random mazes with rectangular-shaped obstacles, similar to the example in Figure \ref{fig:bubble_exp}. The reference trajectory is computed by running the differential FMT$^*$ algorithm \cite{ES-LJ-MP:15a}. In the second set, to test the robustness of the algorithm, we consider a scenario where the reference trajectory is computed disregarding the vehicle's dynamics. This could be the case when, to minimize computation time as much as possible, the use of a motion planner is avoided.  Finally, we consider a scenario where a robotic car is modeled according to a more sophisticated bicycle (equivalently, half-car) model. The reference trajectory is computed by running differential FMT$^*$ on the unicycle model \eqref{eq:dyn}-\eqref{eq:con_3}. By leveraging the differential flatness of the bicycle model, the  $\algor$ algorithm is then applied to the trajectory returned by differential FMT$^*$ (computed on a different model). This scenario represents the typical case whereby one seeks to run a motion planner on a simpler model of a vehicle, and then a smoothing algorithm on a more refined model. Furthermore, this scenario shows how to apply the $\algor$ algorithms to vehicle models more general than \eqref{eq:dyn}-\eqref{eq:con_3}. For all scenarios, the algorithm is stopped whenever the traversal time at the current iteration is no longer reduced with respect to the previous iteration. For the bubble generation method, we chose $r_u = 10\ m$ and $r_l = 1\ m$, consistent with the workspace dimensions discussed below.

All numerical experiments were performed on a computer with an Intel(R) Core(TM) i7-3632QM, 2.20GHz processor and 12GB RAM. The $\algor$ algorithm was implemented in Matlab with an interface to FORCES Pro \cite{AD-JJ:14} for elastic stretching and MTSOS  \cite{TP-SB:14} for speed optimization.

\subsection{Random Mazes}
\begin{figure}[h!]
  \centering
    \includegraphics[width=0.3\textwidth]{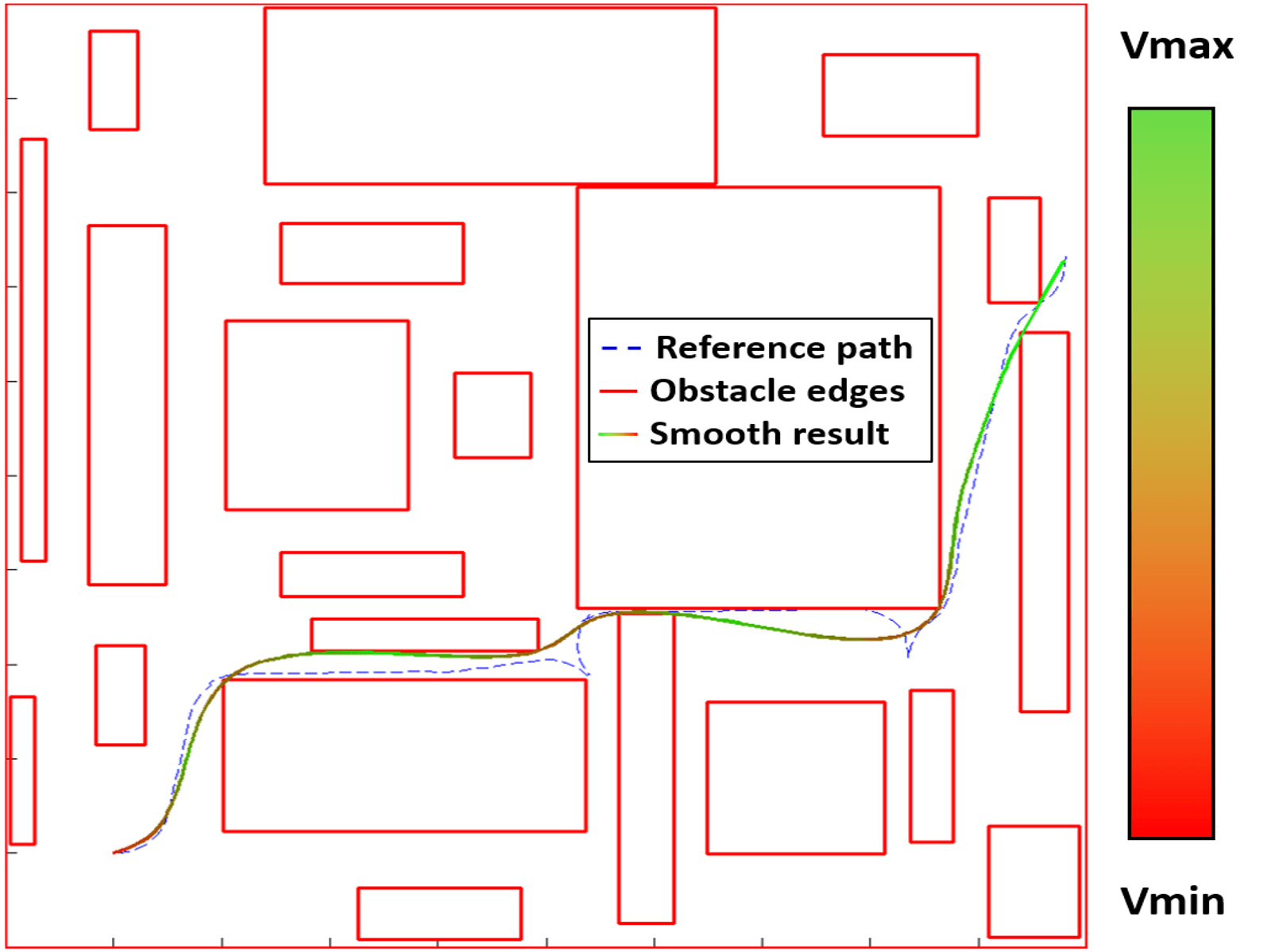}
    \caption{A typical smoothed trajectory for the random maze scenario. In this case, the traversal time reduction is $9.12\%$.}
    \label{fig:random_example}
\end{figure}

In this scenario the workspace is  a $100m \times 100m$ square with rectangular-shaped obstacles randomly placed within (the obstacle coverage was roughly 50\%). The parameters for the model in equations \eqref{eq:dyn}-\eqref{eq:con_3} are $m = 833$ $kg$, $\mu = 0.8$, and $\Ulon = 0.5$ $\mu  m  g$. The reference trajectories, computed via differential FMT$^*$ by using 1,000 samples, were discretized into 257  waypoints with an average segment length equal to $0.56$ $m$.
 On average, each iteration (consisting of bubble generation, shape optimization, and speed optimization) required 119 ms, with a standard deviation of 14 ms. Specifically, the bubble generation algorithm required, on average, 26ms. The shape optimization algorithm required 74 ms. Finally, the speed optimization required 19 ms. A typical smoothed trajectory is portrayed in Figure \ref{fig:random_example}. The traversal time reduction, which is computed according to the formula  $\frac{t_{\mathrm{initial}} - t_{\mathrm{final}}}{t_{\mathrm{initial}}} \cdot 100\%$, ranges from a  minimum of 0.2\% to a maximum of 18\%, with the average value being $3.54\%$. Figure \ref{fig:random_example} shows the smoothed trajectory for one of the 24 random mazes. We note that, apart from the benefit of reduction of traversal time, a smoothed trajectory may be easier to track for a lower-level controller.

\subsection{Lane Changing}
\begin{figure}[h!]
  \centering
    \includegraphics[width=0.5\textwidth]{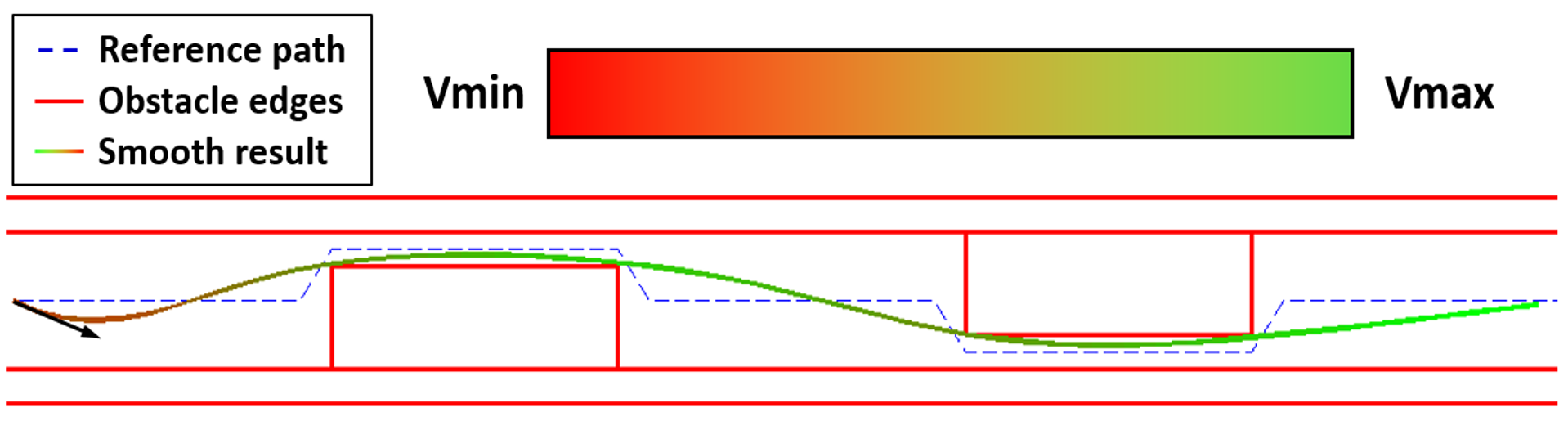}
    \caption{Smoothed trajectory for the lane changing scenario.}
    \label{fig:straightR}
\end{figure}

For this scenario, we consider a road lane 50 $m$ long with rectangular-shaped obstacles in it. The parameters for the model in equations \eqref{eq:dyn}-\eqref{eq:con_3} are $m = 1,725$ $kg$, $\mu = 0.5$, and $\Ulon = 0.3$ $\mu m g$.
The reference trajectory is generated by simply computing the center line of the collision-free ``tube'' along the road. This corresponds to the case where, to minimize computation time as much as possible, a reference trajectory is computed disregarding vehicle's dynamics. Figure \ref{fig:straightR} shows the smoothed trajectory and speed profile. The computation time was $100$ ms. This scenario illustrates  that algorithm  $\algor$ can also smooth reference trajectories that  are not dynamically-feasible. Of course, in this case the traversal time for the smoothed trajectory is  longer, due to the dynamic constraints.

\subsection{Smoothing with Bicycle Model}
\begin{figure}[h!]
  \centering
    \includegraphics[width=0.23\textwidth]{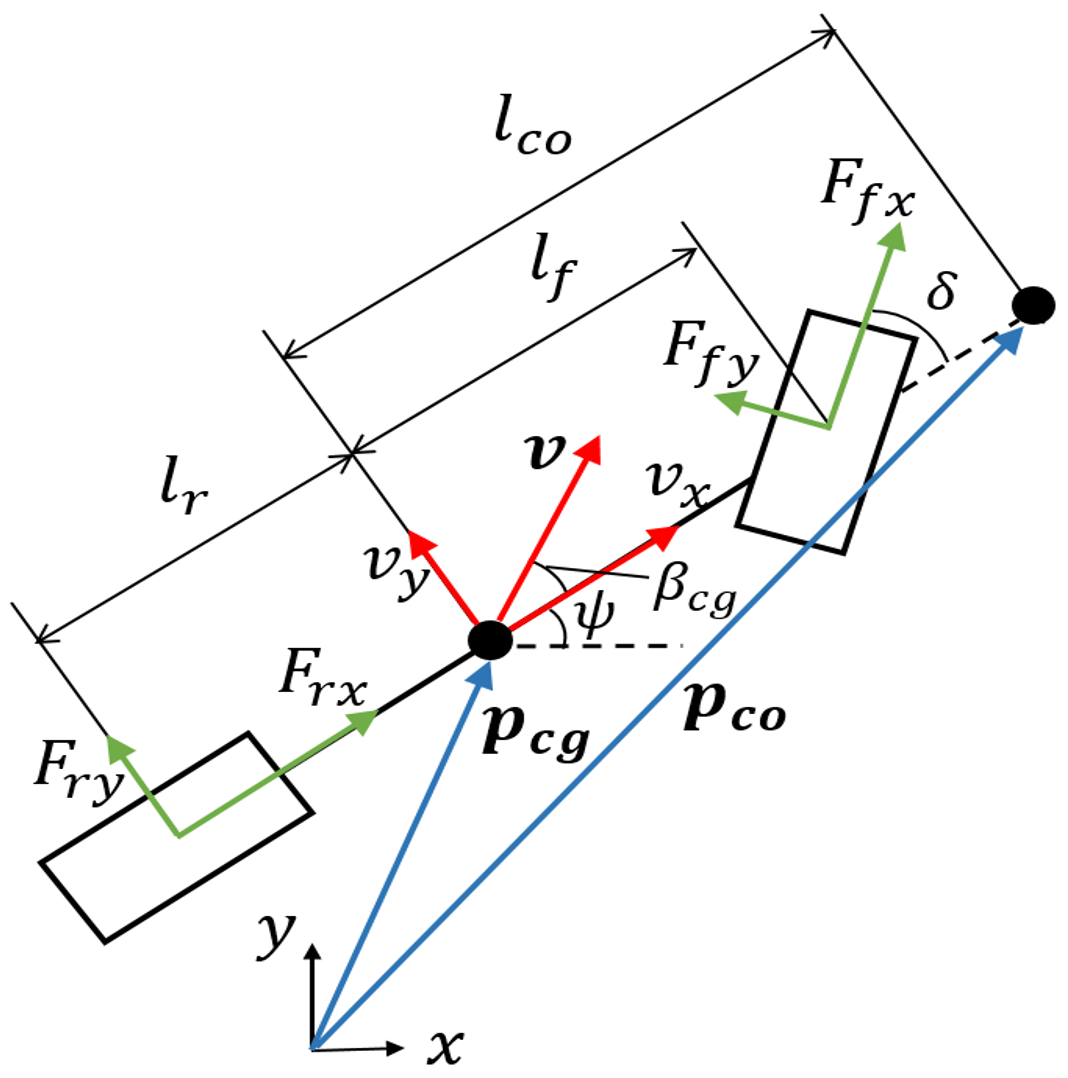}
    \caption{Definition of variables for bicycle model (adapted from \cite{JHJ-RVC-SCP-SK-EF-PT-KI:13}).}
    \label{fig:bike}
\end{figure}

In this scenario, we assume a more sophisticated model for the vehicle, namely a half-car (or bicycle) model, see Figure \ref{fig:bike}. This model is widely used when local vehicle states such as sideslip angle and yaw rate are of primary interest \cite{DLM:03}. In Figure \ref{fig:bike}, $\mathbf{p}_{cg}$ denotes the center of gravity (CG) of the car, $\psi$ denotes vehicle's orientation, and $v_x$ and $v_y$ denote components of speed $\mathbf{v}$  in a body-fixed axis system. Also, $l_f$ and  $l_r$ denote the distances from the CG to the front and rear wheels, respectively. Finally, $F_{\alpha \beta}$, with $\alpha \in \{f,r\}, \beta \in \{x,y\}$ denotes the frictional forces of front and rear wheels. The control input is represented by the triple $\zeta = [\delta,s_{fx},s_{rx}]$, where $\delta$ denotes the steering angle of the front wheel, and $s_{fx}$ and $s_{rx}$ denote the longitudinal slip angles of the front  and rear tires, respectively. See \cite{JHJ-RVC-SCP-SK-EF-PT-KI:13} for more details. Referring to Figure  \ref{fig:bike}, the position $\mathbf{p}_{co}$ can be used as a \emph{differentially flat output} \cite{JA:94}. Specifically, let $m$ be the mass of the vehicle, $I_z$ the yaw moment of inertia, and $l_{co}:=I_z/ml_r$. One can show that
\begin{equation*}
m\ \mathbf{\ddot{p}_{co}} = m\ R(\psi)\begin{bmatrix}
                                    \dot{v_x} - v_y\dot{\psi} - l_{co} \dot {\psi}^2 \\[0.3em]
                                    \dot{v_y} + v_x\dot{\psi} + l_{co}\ddot{\psi}
                                              \end{bmatrix} : = R(\psi)\mathbf{u},
                                              \end{equation*}
where $R(\cdot)$ is the 2D rotation matrix and $\mathbf{u} = [\ulon, \ulat]$ is the flat input comprising longitudinal force $\ulon$ and latitudinal force $\ulat$. Note that the flat dynamics are formally identical to those of the unicycle model \eqref{eq:dyn}-\eqref{eq:con_3}. By leveraging differential flatness, the idea is then to smooth a trajectory in the flat output space and then map the flat input $\mathbf{u}$ to the input $\zeta = [\delta,s_{fx},s_{rx}]$. Constraints for the flat input $\mathbf{u}$ take the same form as in equations \eqref{eq:con_1}-\eqref{eq:con_3} -- the details can be found in \cite{JHJ-RVC-SCP-SK-EF-PT-KI:13}.  When mapping $\mathbf{u}$ into $[\delta,s_{fx},s_{rx}]$, under a no-drift assumption, only two real inputs can be uniquely determined, while the third is effectively a ``degree of freedom," see  \cite[Section II]{JHJ-RVC-SCP-SK-EF-PT-KI:13}. In this paper, we consider as degree of freedom the real input $s_{rx}$. Its value is set equal to the solution of an optimization problem aimed at minimizing the tracking error with respect to the trajectory obtained with the flat input $\mathbf{u}$ (the tracking error is due to the no-drift assumption).

\begin{figure}[h!]
  \centering
    \includegraphics[width=0.28\textwidth]{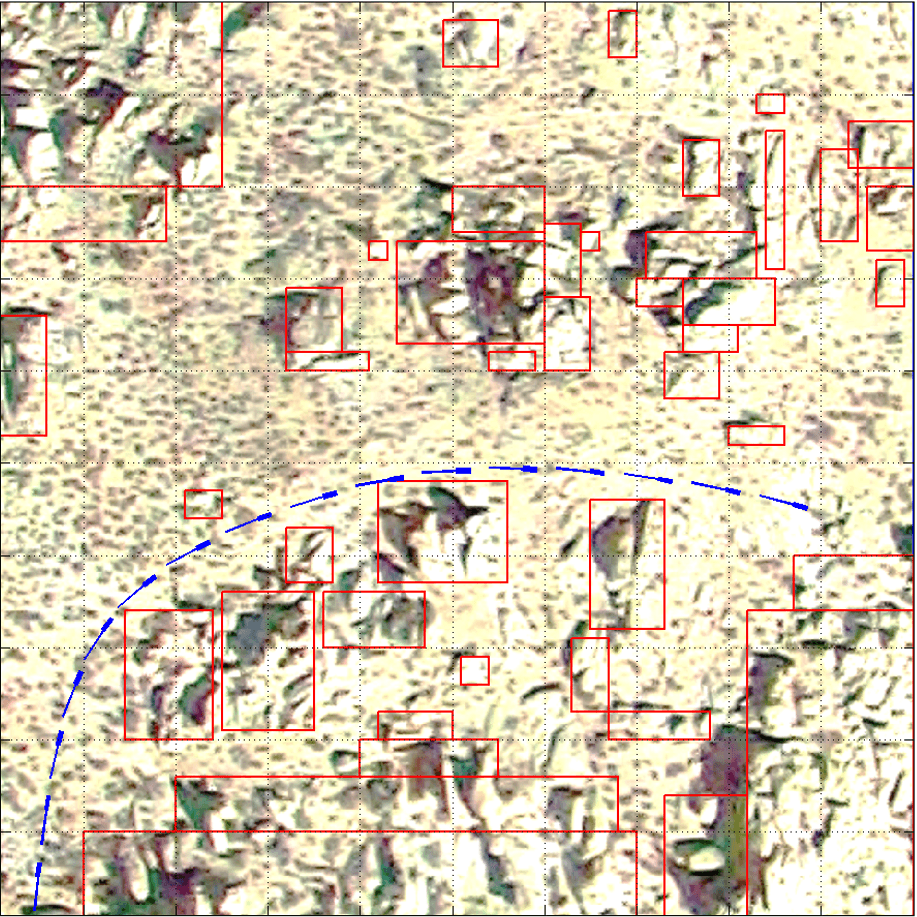}
    \caption{Smoothed trajectory in a rocky terrain with bicycle model (obstacles in work space)}
    \label{fig:rocky3}
\end{figure}

\begin{figure}[h!]
  \centering
    \includegraphics[width=0.3\textwidth]{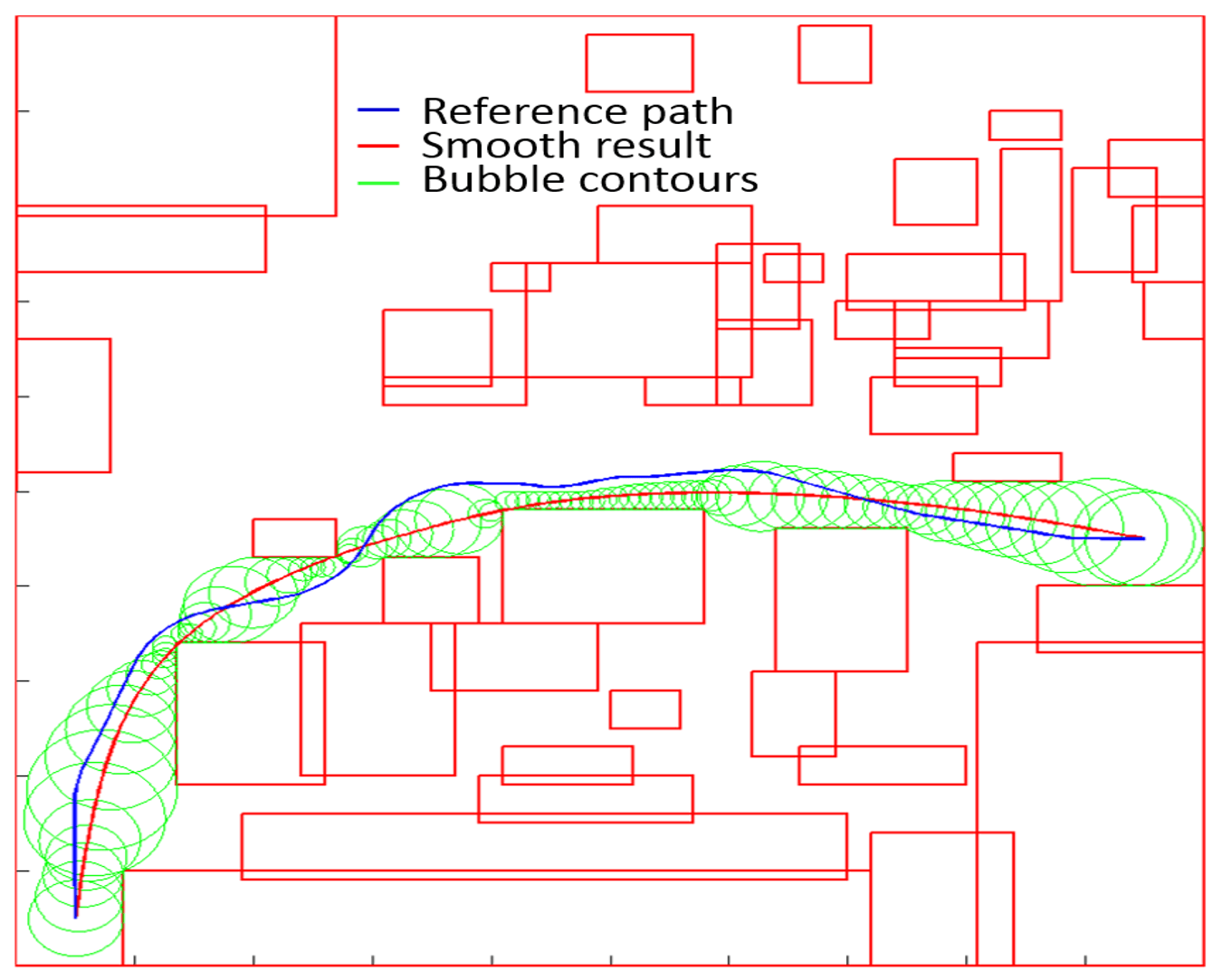}
    \caption{Smoothed trajectory in a rocky terrain with bicycle model  (obstacles inflated in configuration space)}
    \label{fig:rocky1}
\end{figure}

Figure \ref{fig:rocky1} shows the application of the $\algor$ algorithm to a $100m \times 100m$ rocky terrain portrayed in Figure \ref{fig:rocky3}.  The parameters of the bicycle model are $m = 1,725$ $kg$, $I_z = 1,300$ $kg\cdot m^2$, $l_f = 1.35$  $m$, $l_r = 1.15$ $m$, and $h = 0.3$ $m$. In this case, the reference trajectory is computed by running differential FMT$^*$ with 1,000 samples on a unicycle model, as defined in equations \eqref{eq:dyn}-\eqref{eq:con_3}. The reference trajectory is discretized into 257 waypoints. The $\algor$ algorithm is then applied by using the aforementioned bicycle model and differential flatness transformation. Interestingly, in this example the smoothing algorithm is applied to a reference trajectory generated with a different (simpler) model -- this, again, shows the robustness of the proposed approach. The length reduction was 4.19\%, while the traversal time reduction was 39.74\%. Computation times are reported in Table \ref{tab:result_rocky}. Considering two iterations, the total smoothing time is 798 ms. Specifically, about 30\% of the time is taken by differential FMT$^*$, 20\% by the shape optimization algorithm, and the remaining time by the bubble generation, the speed optimization, and the mapping via differential flatness to a bicycle model. Remarkably, this result appears compatible with the real-time requirements of autonomous driving. Indeed, we note that the example in Figure \ref{fig:rocky1}  is rather extreme in that a very long trajectory is planned amid several obstacles. In practical scenarios (e.g., urban driving), the planning problem may be simpler, implying that the computation times would be even lower.

\subsection{Elastic Stretching Moose Test}
\begin{figure}
\centering
\begin{subfigure}{.25\textwidth}
  \centering
  \includegraphics[width=1\linewidth]{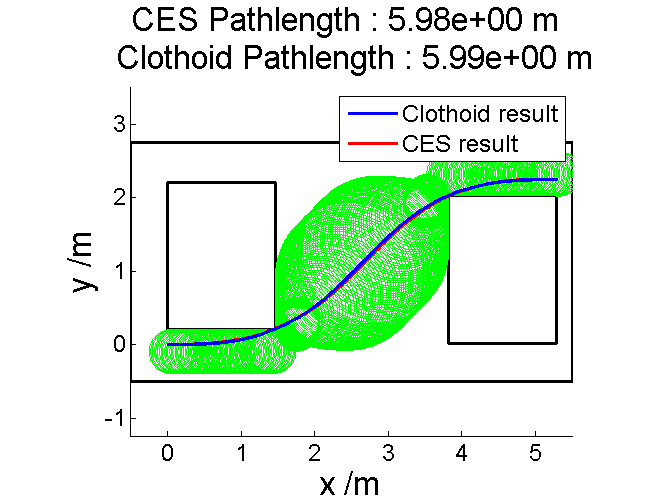}
  \caption{Moose test simulation}
  \label{fig:com_CES_Clothoid}
\end{subfigure}%
\begin{subfigure}{0.23\textwidth}
  \centering
  \includegraphics[width=1\linewidth]{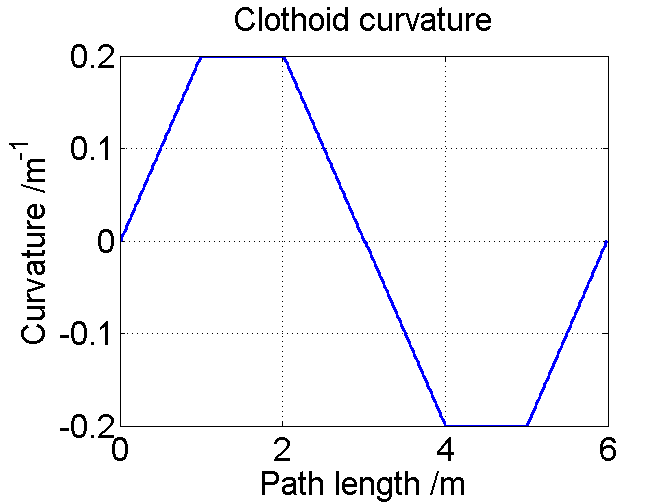}
  \caption{Curvature profile of the clothoid trajectory}
  \label{fig:profile_Clothoid}
\end{subfigure}
\caption{Comparison of CES output and a clothoid trajectory.}
\vspace{-1.0em}
\end{figure}

In order to study in isolation the behavior of the Elastic Stretching algorithm, which is one of the main contributions of this paper, we compare our result with a trajectory consisting of clothoid splines. We note that the objective in the shape optimization step is not simply minimum length (which, for a unicycle model, produces paths consisting only of segments with maximum or zero curvature \cite{LED:57}), but instead encodes a notion of minimum overall curvature more compatible with dynamic considerations and speed optimization. To simplify the problem for finding a optimal solution with clothoid splines, we assume a constant vehicle speed along the trajectory. Fig. \ref{fig:com_CES_Clothoid} illustrates a scenario of a simple Moose test (S shape turn). Here we consider a turning radius lower bound of $5m$, and an upper bound on the path curvature rate of change of $1 m^{-1}s^{-1}$. Fig. \ref{fig:profile_Clothoid} shows the piecewise linear curvature profile for the clothoid trajectory. The results of the Elastic Stretching approach and the clothoid trajectory are very close in this illustrative example, with an error of $0.17\%$ on the total path length.

\subsection{Discussion}
Overall, the above numerical experiments show three major trends. First, the smoothed trajectory often results in a noticeable length and traversal time reduction and, in general, a sequence of waypoints that may be easier to track for a lower-level controller. Second, the $\algor$ algorithm appears robust with respect to the model used to generate the initial reference trajectory. This is a fundamental property, as in practice one would use a motion  planner on a simpler model (e.g., unicycle), and then run a smoothing algorithm with a more sophisticated model. Third, computation times are consistently below one second and in general appear compatible with a real-time implementation.

\begin{table}
\begin{center}
    \begin{tabular}{ | c | c | }
    \hline
                      & \textbf{Time [ms]} \\\hline
    \textbf{Global Planner} &    $357$      \\\hline
    \textbf{Bubble Generation} &    $93$ (2 iterations)      \\\hline
    \textbf{Shape Optimization} &  $203$ (2 iterations)  \\\hline
    \textbf{Speed Optimization} &  $31$ (2 iterations)  \\\hline
    \textbf{Mapping to Half-Car Model} &  $114$ \\\hline
    \textbf{Total} &  $798$ \\\hline
    \end{tabular}
\caption{Computation times for planning and smoothing with a bicycle model.}
\vspace{-.5em}
\label{tab:result_rocky}
\end{center}
\end{table}

\vspace{-.5em}
\section{Conclusions}\label{sec:conc}
In this paper we presented a novel algorithm, $\mathrm{\texttt{Convex}}$ $\mathrm{\texttt{Elastic Smoothing}}$, for trajectory smoothing which  alternates between shape and speed optimization. We showed that both optimization problems can be solved via convex programming, which makes $\algor$ particularly fast and amenable to a real-time implementation.

This paper leaves numerous important extensions open for further research. First, it is of interest to extend the  $\algor$ algorithm to other dynamic systems, such as aerial vehicles or spacecraft. Second, we plan to investigate more thoroughly (1) the robustness of the algorithm when the reference trajectory is not collision-free or dynamically-feasible and (2) the ``typical" factor of suboptimality for a number of representative scenarios. Third, for shape optimization, this paper considered smoothness as the objective function. It is of interest to consider alternative objectives, which, for example, could reproduce the trajectories performed by race car drivers (such trajectories may involve significant curvature variations). Fourth, in an effort to make the proposed algorithm ``trustworthy," we plan to characterize upper bounds for computation times under suitable assumptions on the obstacle space. Finally, we plan to deploy the $\algor$ algorithm on real self-driving cars.

\section*{Acknowledgement}
The authors gratefully acknowledge insightful comments from Ben Hockman and Rick Zhang. This research was supported by an Early Career Faculty grant from
NASA's Space Technology Research Grants Program, grant NNX12AQ43G.

\let\c\originalc
\let\v\originalv
\bibliographystyle{IEEEtran}
\bibliography{../../../bib/alias,../../../bib/main}
\end{document}